\documentclass[review]{elsarticle}
\usepackage{lineno,hyperref}
\usepackage{times}
\usepackage{balance}
\usepackage{amsmath}
\usepackage[tight,footnotesize]{subfigure}
\usepackage{graphicx}
\usepackage{bm}
\usepackage{algorithm}
\usepackage[noend]{algorithmic}
\usepackage{multirow}
\usepackage{slashbox}
\usepackage{caption}
\usepackage{amsfonts}
\newtheorem{theorem}{Theorem}

\newtheorem{lemma}{Lemma}

\modulolinenumbers[5]

\journal{Journal of \LaTeX\ Templates}

\begin{document}

\begin{frontmatter}

\title{Adaptive Robust Optimization with Nearly
Submodular Structure}
\tnotetext[mytitlenote]{Fully documented templates are available in the elsarticle package on \href{http://www.ctan.org/tex-archive/macros/latex/contrib/elsarticle}{CTAN}.}

\author{Shaojie Tang}
\address{Naveen Jindal School of Management, University of Texas at Dallas}
\fntext[myfootnote]{Since 1880.}

\author{Jing Yuan}
\address{Department of Computer Science, University of Texas at Dallas}


%\KEYWORDS{approximation algorithm; team formation; cover decomposition}

\begin{abstract}
Constrained submodular maximization has been extensively studied in the recent years. In this paper, we study adaptive robust optimization with nearly submodular structure (ARONSS). Our objective is to randomly select a subset of items that maximizes the worst-case value of several reward functions simultaneously. Our work differs from existing studies in two ways: (1) we study the robust optimization problem under the adaptive setting, i.e., one needs to adaptively select items based on the feedback collected from picked items, and (2) our results apply to a broad range of reward functions characterized by $\epsilon$-nearly submodular function. We first analyze the adaptvity gap of ARONSS and show that the gap between the best adaptive solution and the best non-adaptive solution is bounded. Then we propose a approximate solution to this problem when all reward functions are submodular. Our algorithm achieves approximation ratio $(1-1/e)$ when considering matroid constraint.  At last, we present two heuristics for the general case. All proposed solutions are non-adaptive which are easy to implement.
\end{abstract}

%\begin{keyword}
%\texttt{elsarticle.cls}\sep \LaTeX\sep Elsevier \sep template
%\MSC[2010] 00-01\sep  99-00
%\end{keyword}

\end{frontmatter}
%%%%%%%%%%%%%%%%%%%%%%%%%%%%%%%%%%%%%%%%%%%%%%%%%%%%%%%%%%%%%%%%%%%%%%

% Samples of sectioning (and labeling) in IJOC
% NOTE: (1) \section and \subsection do NOT end with a period
%       (2) \subsubsection and lower need end punctuation
%       (3) capitalization is as shown (title style).
%
%\section{Introduction.}\label{intro} %%1.
%\subsection{Duality and the Classical EOQ Problem.}\label{class-EOQ} %% 1.1.
%\subsection{Outline.}\label{outline1} %% 1.2.
%\subsubsection{Cyclic Schedules for the General Deterministic SMDP.}
%  \label{cyclic-schedules} %% 1.2.1
%\section{Problem Description.}\label{problemdescription} %% 2.

\section{Introduction}\label{sec:introduction}
Constrained submodular maximization has attracted growth attention recently \cite{calinescu2011maximizing}\cite{buchbinder2018deterministic}\cite{buchbinder2014submodular}. Most existing work on submodular maximization focus on selecting a subset of items subject to given constraints so as to maximize a submodular objective function \cite{krause2008near}.  In this paper, we study adaptive robust optimization with nearly submodular structure (ARONSS). This study belongs to the category of robust submodular maximization. Our objective is to randomly select a subset of items that performs well over  several reward functions. Although robust submodular maximization has been well studied \cite{anari2017structured}\cite{krause2008robust}\cite{chen2016robust}\cite{orlin2018robust}, most of existing studies assume an non-adaptive setting, i.e., one has to select a subset of items all at once in advance, and submodular reward function. However, in many applications from artificial intelligence, the outcome of an objective function is often uncertain, one needs to make a sequence of decisions adaptively based on the outcomes of the previous decisions  \cite{golovin2011adaptive}.  Moreover, the reward function is not necessarily submodular. This motivates us to study the adaptive robust optimization problem with general reward functions.

The main contribution of this paper is three-fold:
\begin{itemize}
\item We extend the previous studies on robust submodular maximization in two directions: (1) we consider the robust optimization problem under the adaptive setting, i.e., one can select one item at a time and observe the outcome of picked items, before selecting the next item, and (2) our results apply to a broad range of reward functions characterized by $\epsilon$-nearly submodular function.
\item We first analyze the adaptivity gap of ARONSS and show that the gap between the best adaptive solution and the best non-adaptive solution is bounded. This  enables us to focus on designing non-adaptive solutions which are much easier to work with.
\item  Then we propose an approximate solution to this problem when all reward functions are submodular. The approximation ratio is $(1-1/e)$ when considering matroid constraint. We also present two algorithms that achieve bounded approximation ratios for the general case. All algorithms are non-adaptive and easy to implement.
\end{itemize}

\section{Preliminaries and Problem Formulation}
\label{sec:system}
\subsection{Submodular Function} A set function $h(S)$ that maps subsets of a finite ground set $\Omega$ to non-negative real numbers is said to be submodular  if for every $S_1, S_2 \subseteq \Omega$ with $S_1 \subseteq S_2$ and every $v \in \Omega \backslash S_2$, we have that \[h(S_1\cup \{v\})-h(S_1)\geq h(S_2\cup \{v\})-h(S_2)\]
A submodular function $h$ is said to be monotone if $h(S_1) \leq h(S_2)$ whenever $S_1 \subseteq S_2$.
\subsection{Items and States} Let $E$ denote a finite set of $n$ items, and each item $e\in E$ is in a particular state from a set $O$ of possible states.  Let $\phi: E\rightarrow O$ denote a realization of item states.   Each item $e$ is associated with a random variable $Y_e$ that represents a random realization of $e$'s state. We use $\mathbf{Y}_E=\{Y_e\mid e\in E\}$ to denote the collection of all variables. We assume there is a known prior probability distribution $\mathcal{D}_e$ over realizations for each item $e$, i.e., $\mathcal{D}_e=\{\Pr[Y_e =y_e]: y_e\in O\}$. We further assume that the states of all items are decided independently from each other, i.e., $\mathbf{Y}_E$ is drawn randomly from the product distribution $\prod_{e\in E}\mathcal{D}_e$. We use  $\mathbf{y}_E=\{y_e\mid e\in E\}$ to denote the realization of items' states. After picking an item $e$, we are able to observe its state $Y_e=y_e$.
%\subsection{Objective Function}
%For any $Z\subseteq E$, we say a partial realization $\psi_Z$ is feasible if $\Pr[\Psi_Z=\psi_Z]> 0$, i.e., $\psi_Z$ is observed with positive probability. We use $O_Z \subseteq  Z\times O$ to denote the entire set of feasible realizations of $Z$.
\subsection{$\epsilon$-nearly Submodular Reward Functions}
 We are given a family of reward functions $\mathcal{F}=\{f_1, f_2, \cdots, f_m\}$, where each $f_i\in \mathcal{F}: 2^{E\times O}\rightarrow \mathbb{R}_{\geq 0}$ maps a set of items and  their states $X \subseteq E\times O$ to some reward $\mathbb{R}_{\geq 0}$. In this work, we assume each function $f_i$ is monotone, i.e., $f_i(A)\leq f_i(B)$ for all $A \subseteq B$,  and \emph{$\epsilon$-nearly submodular}, i.e., for any $f_i\in \mathcal{F}$, there is a submodular function $g_i$ such that for any $X \subseteq E\times O$, we have $\epsilon g_i(X) \leq f_i(X) \leq \frac{1}{\epsilon}g_i(X)$ where $\epsilon\in(0,1]$. It is easy to verify that any submodular function is $1$-nearly submodular. %Define $f_i_B(A):=f_i(B\cup A)-f_i(B)$ as the marginal value of $A$ with respect to $B$.
 \subsection{Adaptive Policies} We model the adaptive strategy of picking items through a policy $\pi$ \cite{golovin2011adaptive}. Formally, a policy $\pi$ is a function that specifies which item to pick next under the  observations made so far: $\pi: 2^{V\times O}\rightarrow E$. Note that $\pi$ can be regarded as some decision tree that specifies a rule for  picking items adaptively.  Assume that when the items are in state $\mathbf{Y}_E=\mathbf{y}_E$, the policy $\pi$ picks a set of items (and corresponding states), which is denoted by $S(\pi, \mathbf{y}_E)\subseteq E\times O$.
Thus, given the policy $\pi$, its expected reward received from function $f_i$ is $\mathcal{U}(\pi, f_i):=\mathbb{E}_{\mathbf{y}_E}[f_i(S(\pi, \mathbf{y}_E))]$.
  In the context of robust optimization, our goal is to pick a set of items (and corresponding states) that achieves high reward in the worst-case over reward functions in $F$. Thus, we define the utility $\mathcal{U}(\pi, \mathcal{F})$ of  $\pi$ as
\[\mathcal{U}(\pi, \mathcal{F})=\min_{i\in [m]}\mathcal{U}(\pi, f_i)\]

 Let $\mathcal{I}$ be a  \emph{downward-closed} family of subsets of $E$, i.e., a family of subsets $\mathcal{I}$ is downward-closed if for any subset in $\mathcal{I}$, it also belongs to $\mathcal{I}$. We use $E(\pi, \mathbf{y}_E)$ to refer to the subset of items picked by policy $\pi$ given state $\mathbf{y}_E$. We say a policy $\pi$ is \emph{feasible} if for any $\mathbf{y}_E$, $E(\pi, \mathbf{y}_E)\in \mathcal{I}$. This  downward-closed family generalizes many useful constraints such as  matroid and knapsack constraints. Our goal is to identify the best feasible policy that maximizes its expected utility.
\[\max_{\pi} \mathcal{U}(\pi, \mathcal{F}) \mbox{ subject to $E(\pi, \mathbf{y}_E)\in \mathcal{I}$ for any $\mathbf{y}_E$.}\]

% Given any partial realizations $\psi_U \in U\times O$ and $\psi_V \in V\times O$ of two disjoint sets of items $U$ and $V$, we use $\Pr[\psi_U|\psi_V]$ to denote the probability of $\psi_U$ being observed conditioned on $\psi_V$ are observed. %In case $\psi_V$ is not a feasible realization (i.e., the probability that $\psi_V$ is observed is zero), we set $\Pr[\psi_U|\psi_V]=0$ for all $\psi_U \in U\times O$.
%\begin{definition} [Degree of Dependence] The degree of dependence of  a known prior probability distribution $\mathcal{D}$ is defined as follows.
%\begin{equation}
%\gamma(\mathcal{D}):=\sup \{c \mid\forall V\subseteq E, U\subseteq E\setminus V, \psi_U\in O_U, \psi_V\in O_V,\psi'_V\in O_V: \Pr[\psi_U|\psi_V]\leq c \Pr[\psi_U|\psi'_V]\}
%\end{equation}
%\end{definition}

%Degree of dependence refers to the degree to which one group of items' realizations could affect the other group of items' realizations. If these realizations are assigned independently across different items, then the degree of dependence is 1.
\section{Analysis on Adaptivity Gap}
\label{sec:4}
We say a policy is non-adaptive if it always picks the next item independent of the states of the picked items. Clearly adaptive polices obtain at
least as much utility as non-adaptive policies. Perhaps surprisingly,  building on recent advances in stochastic submodular probing \cite{bradac2019near}, we show that this adaptivity gap is upper bounded by a constant (given that $\epsilon$ is a constant). Based on this result, we can focus on designing non-adaptive polices which are much easier to work with.
\begin{theorem}
\label{thm:11}
Given any adaptive policy $\pi$, there exists a non-adaptive algorithm $\sigma_\pi$ such that $\mathcal{U}(\sigma_\pi, \mathcal{F})\geq \frac{\epsilon^2}{2}\mathcal{U}(\pi, \mathcal{F})$.
\end{theorem}
\emph{Proof:} Given any adaptive policy $\pi$, we follow the idea in \cite{gupta2017adaptivity} and define a non-adaptive policy $\sigma_\pi$: randomly draw a state vector $\mathbf{y}_E$ from the product distribution $\prod_{e\in E}\mathcal{D}_e$ (this step is done virtually), pick  $E(\pi, \mathbf{y}_E)\subseteq E$, i.e., pick all items picked by $\pi$ given $\mathbf{y}_E$. %Let $e$ be the first item picked by $\sigma_\pi$ (and $\pi$),
Let $\mathbf{y}'_E$ be the state of all items drawn virtually by $\sigma_\pi$ and $\mathbf{y}_E$ be the true state of all items when picked by $\sigma_\pi$. %Let $\sigma_\pi_A$ (resp. $\sigma_\pi_B$) denote the execution of $\sigma_\pi$ after observing $A$ (resp. $B$) and  $\pi_A$ (resp. $\pi_B$) denote the execution  of $\pi$ after observing $A$ (resp. $B$).

Now consider any $i\in[m]$, the expected value of $f_i$  obtained by $\sigma_\pi$ is
\begin{equation}\mathcal{U}(\sigma_\pi, f_i)=\mathbb{E}_{\mathbf{y'}_E}\left[\mathbb{E}_{\mathbf{y}_E}[f_i(\bigcup_{e\in E(\pi, \mathbf{y}'_E)}(e, y_e))]\right]\label{eq:1}
\end{equation}
Because $f_i$ is $\epsilon$-nearly submodular, we have
\begin{equation}
\mathbb{E}_{\mathbf{y'}_E}\left[\mathbb{E}_{\mathbf{y}_E}[f_i(\bigcup_{e\in E(\pi, \mathbf{y}'_E)}(e, y_e))]\right]\geq \mathbb{E}_{\mathbf{y'}_E}\left[\mathbb{E}_{\mathbf{y}_E}[\epsilon g_i(\bigcup_{e\in E(\pi, \mathbf{y}'_E)}(e, y_e))]\right]=\epsilon \mathcal{U}(\sigma_\pi, g_i) \label{eq:2}
\end{equation}
(\ref{eq:1}) and (\ref{eq:2}) together imply that
\begin{equation}
\mathcal{U}(\sigma_\pi, f_i)\geq\epsilon \mathcal{U}(\sigma_\pi, g_i)\label{eq:3}
\end{equation}
We next analyze the utility of $\pi$. The expected value of $f_i$ obtained by $\pi$ is
\begin{equation}
\mathcal{U}(\pi, f_i)=\mathbb{E}_{\mathbf{y}_E}[f_i(S(\pi, \mathbf{y}_E))]
\label{eq:3}\end{equation}
Because $f_i$ is $\epsilon$-nearly submodular, we have
\begin{equation}\mathbb{E}_{\mathbf{y}_E}[f_i(S(\pi, \mathbf{y}_E))]\leq \mathbb{E}_{\mathbf{y}_E}[\frac{1}{\epsilon}g_i(S(\pi, \mathbf{y}_E))]=\frac{1}{\epsilon}\mathcal{U}(\pi, g_i)
\label{eq:4}
\end{equation}
(\ref{eq:3}) and (\ref{eq:4}) together imply that
\begin{equation}\mathcal{U}(\pi, f_i) \leq \frac{1}{\epsilon}\mathcal{U}(\pi, g_i)\label{eq:6}\end{equation}
Because $g_i$ is submodular, the ratio between $\mathcal{U}(\pi, g_i)$ and $\mathcal{U}(\sigma_\pi, g_i)$ is upper bounded by $2$ \cite{bradac2019near}, i.e., $\mathcal{U}(\pi, g_i)\leq2 \mathcal{U}(\sigma_\pi, g_i)$. This together with (\ref{eq:3}) and (\ref{eq:6}) imply that
\begin{equation}
\mathcal{U}(\sigma_\pi, f_i) \geq \frac{\epsilon^2}{2}  \mathcal{U}(\pi, f_i)
\end{equation}
It follows that
\begin{align}\mathcal{U}(\sigma_\pi, \mathcal{F})&=\min_{i\in [m]}\mathcal{U}(\sigma_\pi, f_i)\\
 &\geq \min_{i\in [m]}\frac{\epsilon^2}{2}  \mathcal{U}(\pi, f_i)\\
 &= \frac{\epsilon^2}{2} \mathcal{U}(\pi, \mathcal{F})
\end{align} $\Box$

It was worth noting that Theorem \ref{thm:11} holds when $\mathcal{I}$ is a \emph{prefix-closed} family of constraints, i.e., a family of subsets $\mathcal{I}$ is prefix-closed if for any subsequence in $\mathcal{I}$, its prefix also belongs to $\mathcal{I}$.
\begin{algorithm}[hpt]
{\small
\caption{$\sigma^{\mathrm{1/m}}$}
\label{alg:greedy-peak}
%\textbf{Input:} Social network $\mathcal{G}$, budget $\mathcal{B}$, individual attention constraint $\kappa_i$, overall attention constraint $K$.\\
%\textbf{Output:} Seed set $S$.
\begin{algorithmic}[1]
\STATE Set $i=1$.
\WHILE{$i\leq m$}
\STATE $E_i\leftarrow \mathrm{APPROX}(\max_{S\in \mathcal{I}}\mathcal{U}(S, f_i))$
\STATE $i\leftarrow i+1$
\ENDWHILE
\STATE Randomly pick an index $i\in [m]$
\RETURN $E_i$
\end{algorithmic}
}
\end{algorithm}
\section{Approximate Solution for Submodular Reward Function}
We first focus on the case when $\epsilon=1$, i.e., all reward functions are submodular. We propose a constant approximate solution to this special case. The basic idea of our approach is that we first derive a constant approximate solution to the non-adaptive robust optimization problem and Theorem \ref{thm:11} implies that this solution is also a constant approximate solution to the original problem.

We first introduce the non-adaptive robust optimization problem with submodular structure. Given any reward function $f_i$, we use $f_i(V)$ to denote the expected reward of selecting $V\subseteq E$. Given a non-adaptive policy $\sigma$, let $\mathcal{U}(\sigma, f_i):=\sum_{V\in \mathcal{I}}\beta^\sigma_V f_i(V)$ denote the expected reward gained from function $f_i$ where $\beta^\sigma_V $ is the probability that $V$ is selected by $\sigma$. The utility $\mathcal{U}(\sigma, \mathcal{F})$ of $\sigma$ is $\mathcal{U}(\sigma, \mathcal{F})=\min_{i\in[m]}\mathcal{U}(\sigma, f_i)$.  We next formulate the non-adaptive robust optimization problem as follows.
%\[\max_{\sigma} \mathcal{U}(\sigma, f_i) \mbox{ subject to $\sum_{V\in \mathcal{I}}\beta^\sigma_V\leq 1$ and for any $U\notin\mathcal{I}, \beta^\sigma_U=0$.}\]

\begin{center}
\framebox[0.6\textwidth][c]{
\enspace
\begin{minipage}[t]{0.6\textwidth}
\small
\textbf{P.1} $\max_{\sigma} \mathcal{U}(\sigma, \mathcal{F})$\\
\textbf{subject to:}
\begin{equation*}
\begin{cases}
\mathcal{U}(\sigma, \mathcal{F})=\min_{i\in[m]}\mathcal{U}(\sigma, f_i)\\
\forall i\in[m], \mathcal{U}(\sigma, f_i):=\sum_{V\in \mathcal{I}}\beta^\sigma_V f_i(V)\\
\sum_{V\in \mathcal{I}}\beta^\sigma_V\leq 1
\end{cases}
\end{equation*}
\end{minipage}
}
\end{center}
\vspace{0.1in}

Before introducing our algorithm, we first introduce some important notations.  For a independence system $\mathcal{I}$, the polytope of $\mathcal{I}$ is defined as
$P(\mathcal{I}) = \mathrm{conv}\{\mathbf{1}_I : I \in \mathcal{I}\}$ where $\mathbf{1}_I\in[0,1]^n$ denotes the vector with entries $I$ one and all other entries zero. Given a vector $\mathbf{x}\in[0,1]^n$, the multilinear extension of $f$ is defined as $F(\mathbf{x})=\sum_{X\subseteq \Omega} f(X) \prod_{i\in X} x_i \prod_{i\notin X} (1-x_i)$. Define the marginal of $e$ for $F$ as $F(e|\mathbf{x})=F(\mathbf{x}\vee \mathbf{1}_{e})-F(\mathbf{x})$ where $\mathbf{x}\vee \mathbf{1}_{e} $ denotes the component wise maximum.

As a corollary of Theorem \ref{thm:11}, i.e.,  when $\epsilon=1$, the following lemma bounds the adaptivity gap when all reward functions are submodular.
\begin{lemma}
\label{lem:1}
Let  $\pi^{*}$ denote the optimal adaptive policy and $\sigma^*$ denote the optimal non-adaptive policy, we have $\mathcal{U}(\sigma^{*}, \mathcal{F})\geq \frac{1}{2}\mathcal{U}(\pi^{*}, \mathcal{F})$.
\end{lemma}

We next propose a continuous greedy algorithm that achieves a constant approximation ratio of \textbf{P.1}. We follow the framework of \cite{chekuri2010dependent} to derive the following lemma.

\begin{lemma}
\label{lem:2}
Given $m$ submodular functions $f_i$ and a value $\gamma$, independence system $\mathcal{I}$, the continuous greedy algorithm finds a point $\mathbf{x}(T)\in P(\mathcal{I})$ such that $F_i(\mathbf{x}(T))\geq  (1 - 1/e)\gamma, \forall i$ or outputs a certificate that there is solution with $F_i(\mathbf{x}(T))\geq \gamma, \forall i$.
\end{lemma}
 \emph{Proof:}  Consider any vector $\mathbf{x}$. If there exists policy, say $\sigma'$, such that $\mathcal{U}(\sigma', f_i) \geq \gamma, \forall i$, we have
  \begin{eqnarray}
\gamma \leq \mathcal{U}(\sigma', f_i)=\sum_{V\in \mathcal{I}}\beta^{\sigma'}_V f_i(V) &\leq&  \sum_{V\in \mathcal{I}}\beta^{\sigma'}_V(F(\mathbf{x}) + \sum_{e\in V} F(e|\mathbf{x}))\nonumber\\
 % &\leq&  \sum_{\phi\in \mathcal{U}} \beta_\phi (F(\mathbf{x}) + \sum_{e\in E(\pi^\diamond, \phi)} F_{\mathbf{x} \setminus e}(\phi_{e})) \nonumber\\
%  &\leq& F(\mathbf{x})+  \sum_{\phi\in \mathcal{U}}\sum_{e\in E(\pi^\diamond, \phi)} (\beta_\phi F_{\mathbf{x}\setminus e}(\phi_{e}))\nonumber\\
  %&\leq& f(H)+  \sum_{\phi\in \Phi}\sum_{v\in V(\Phi)} (\alpha_Y f_{H\setminus e}(\phi_{e}))\\
  &=& F(\mathbf{x})+  \sum_{e\in E}(\sum_{V\in \mathcal{I}\wedge e\in V}\beta^{\sigma'}_V) F(e|\mathbf{x}) \label{line:3}
   \end{eqnarray}
In other words, for any fractional solution $\mathbf{x}$, there exits a direction $v^*(\mathbf{x})\in P(\emph{I})$ where the entry of $e$ is $v^*(\mathbf{x})(e)=\sum_{V\in \mathcal{I}\wedge e\in V}\beta^{\sigma'}_V$ such that  $v^*(\mathbf{x}) \cdot \nabla F(\mathbf{x})\geq \gamma-F(\mathbf{x}), \forall i$. And this direction can be found using linear program. We follow the continuous greedy algorithm and obtain a solution $\mathbf{x}(T)$  such that $F_i(\mathbf{x}(T))\geq  (1 - 1/e)\gamma, \forall i$.

If such policy does not exist, we output a certificate that there is feasible solution that achieves utility $\gamma$. $\Box$

Based on Lemma \ref{lem:2}, we can perform a binary search on $\gamma$ to find a $(1-1/e)$-approximate fractional solution. At last, depending on the type of $\mathcal{I}$, we use an appropriate technique to round the fractional solution to an integral solution. Lemma \ref{lem:1} and Lemma \ref{lem:2} imply the following main result.

\begin{theorem}
\label{them:2}
Our algorithm returns a solution that achieves approximation ratio $\frac{1}{2}(1-1/e)\zeta$ where $\zeta\in[0,1]$ is the performance loss due to rounding.
\end{theorem}

Note that when the constraint is a matroid, we can use swap rounding \cite{chekuri2010dependent} to achieve $\zeta=1$. Many other useful constraints such as knapsack and the intersection of knapsack and matroid constraints admit good rounding techniques \cite{chekuri2014submodular}.

\section{Two Heuristics for Nearly Submodular Reward Functions}
In this section, we introduce two algorithms for computing approximate solutions for the general case. Since the adaptivity gap is bounded in Section \ref{sec:4}, we focus on building non-adaptive policies. In the rest of this paper, we use $\sigma$ to denote a non-adaptive policy.
\subsection{A $1/m$-approximate Solution}
 The basic idea of the first algorithm $\sigma^{\mathrm{1/m}}$ (Algorithm \ref{alg:greedy-peak}) is very simple, we first solve $\max_{\sigma}\mathcal{U}(\sigma, f_i)$ for each $i\in [m]$, then randomly pick one among $m$ outputs as solution. Since we focus on designing non-adaptive solutions, for notation convenience, define $\mathcal{U}(S, f_i)$ as the expected value of $f_i$ obtained from picking $S\subseteq E$ (irrespective of items' states), i.e., $\mathcal{U}(S, f_i)=\mathbb{E}_{\mathbf{y}_E}\left[f_i(\bigcup_{e\in S}(e, y_e))\right]$. One can verify that solving  $\max_{\sigma}\mathcal{U}(\sigma, f_i)$ is equivalent to solving
 $\max_{S\in \mathcal{I}}\mathcal{U}(S, f_i)$.

 To carry out these steps, $\sigma^{\mathrm{1/m}}$ requires one oracle $\mathrm{APPROX}(\max_{S\in \mathcal{I}}\mathcal{U}(S, f_i))$ which returns an approximate solution to $\max_{S\in \mathcal{I}}\mathcal{U}(S, f_i)$ for each $i\in[m]$. Assume the approximation ratio of $\mathrm{APPROX}(\max_{S\in \mathcal{I}}\mathcal{U}(S, f_i))$ is $\alpha_i$, we have
\begin{theorem}
Assume $\pi^*$ is the optimal adaptive policy and $\alpha=\min_{i\in[m]} \alpha_i$, our first policy $\sigma^{\mathrm{1/m}}$ achieves  $ \frac{\epsilon^2 \alpha }{2m}$ approximation ratio for  ARONSS, i.e., $\mathcal{U}(\sigma^{\mathrm{1/m}}, \mathcal{F}) \geq \frac{\epsilon^2 \alpha }{2m}\mathcal{U}(\pi^*, \mathcal{F})$. The time complexity of $\sigma^{\mathrm{1/m}}$ is $O(m\delta)$ where $\delta$ is the time complexity of $\mathrm{APPROX}$.
\end{theorem}
\emph{Proof:} First, according to the definition of $f$, for any $i\in [m]$, we have
\begin{equation}
\label{eq:999}
\max_{S\in \mathcal{I}}\mathcal{U}(S, f_i)=\max_{\sigma}\mathcal{U}(\sigma, f_i) \geq\max_{\sigma} \min_{i\in [m]}\mathcal{U}(\sigma, f_i) =\max_{\sigma}\mathcal{U}(\sigma, \mathcal{F})
 \end{equation}

 Based on the design of $\sigma^{\mathrm{1/m}}$, $\mathrm{APPROX}(\max_{S\in \mathcal{I}}\mathcal{U}(S, f_i))$ is returned as the final solution with probability $1/m$. Because $\mathrm{APPROX}(\max_{S\in \mathcal{I}}\mathcal{U}(S, f_i))$ achieves approximation ratio $\alpha$, we have $\mathcal{U}(\sigma^{\mathrm{1/m}}, f_i)\geq \frac{\alpha}{m}\max_{S\in \mathcal{I}}\mathcal{U}(S, f_i)$, it follows that $\mathcal{U}(\sigma^{\mathrm{1/m}}, f_i)\geq \frac{\alpha}{m}\max_{\sigma}\mathcal{U}(\sigma, \mathcal{F})$. Thus,
 \[\mathcal{U}(\sigma^{\mathrm{1/m}}, \mathcal{F})=\min_{i\in [m]}\mathcal{U}(\sigma^{\mathrm{1/m}}, f_i)\geq \frac{\alpha}{m}\max_{\sigma}\mathcal{U}(\sigma, \mathcal{F})\] due to (\ref{eq:999}). Since $\max_{\sigma}\mathcal{U}(\sigma, \mathcal{F})\geq \frac{\epsilon^2}{2}\mathcal{U}(\pi^*, \mathcal{F})$ due to Theorem \ref{thm:11}, we have $\mathcal{U}(\sigma^{\mathrm{1/m}}, \mathcal{F}) \geq \frac{\epsilon^2 \alpha }{2m}\mathcal{U}(\pi^*, \mathcal{F})$. This finishes the proof of the first part of this theorem. The proof of time complexity is trivial since $\sigma^{\mathrm{1/m}}$  calls $\mathrm{APPROX}$  $m$ times. $\Box$

\paragraph{Discussion on the value of $\alpha$} We next briefly discuss possible solutions to $\max_{S\in \mathcal{I}}\mathcal{U}(S, f_i)$. Consider a special case when all reward functions in $F$ are submodular, i.e., $\epsilon=1$, and $\mathcal{I}$ is a family of subsets that satisfies a knapsack constraint or a matroid constraint \cite{calinescu2011maximizing}, there exist algorithms that achieve $1-1/e$ approximation ratio, i.e., $\alpha=1-1/e$. For more complicated constraints such as intersection of a fixed number of knapsack and matroid constraints, \cite{chekuri2014submodular} provide approximate solutions via the multilinear relaxation and contention resolution schemes.

\subsection{Double-Oracle Algorithm}
We next present a double-oracle based solution to ARONSS. %One can verify that $\mathcal{U}(S, f_i)$ is also $\epsilon$-nearly submodular, i.e., $\epsilon g_i(S)\leq \mathcal{U}(S, f_i)\leq \frac{1}{\epsilon}g_i(S)$.
 We first introduce an optimization problem \textbf{P.2} as follows.
 \begin{center}
\framebox[0.5\textwidth][c]{
\enspace
\begin{minipage}[t]{0.5\textwidth}
\small
\textbf{P.2:} \emph{Maximize $\min_{i\in[m]}\sum_{S\in \mathcal{I}}x_S \mathcal{U}(S, f_i)$}\\
\textbf{subject to:}\begin{align*}
\begin{cases}
\sum_{S\in \mathcal{I}}x_S=1\\
x_S \geq 0, \mbox{               } \forall S \in \mathcal{I}
\end{cases}
\end{align*}
\end{minipage}
}
\end{center}
\vspace{0.1in}
In \textbf{P.2}, $x_S$ indicates the probability of picking $S$. It is easy to verify that finding $\arg\max_{\sigma}\mathcal{U}(\sigma, \mathcal{F})$ is equivalent to solving \textbf{P.2}. In practice, \textbf{P.2} is often solved by the double oracle algorithm \cite{mcmahan2003planning}. Without loss of generality, assume that double oracle algorithm $\sigma^{\mathrm{DO}}$ finds a $\beta$ approximate solution to \textbf{P.2}, i.e., $\mathcal{U}(\sigma^{\mathrm{DO}}, \mathcal{F}) \geq \beta \max_{\sigma}\mathcal{U}(\sigma, \mathcal{F})$, we have $\mathcal{U}(\sigma^{\mathrm{DO}}, \mathcal{F}) \geq \frac{\epsilon^2\beta}{2}\mathcal{U}(\pi^*, \mathcal{F})$ due to the adaptivity gap proved in Theorem \ref{thm:11}.
\begin{theorem}
\label{thm:1}
Assume $\sigma^{\mathrm{DO}}$ finds a $\beta$ approximate solution to \textbf{P.2}, $\sigma^{\mathrm{DO}}$ achieves $\frac{\epsilon^2\beta}{2}$ approximation ratio for ARONSS, i.e., $\mathcal{U}(\sigma^{\mathrm{DO}}, \mathcal{F}) \geq \frac{\epsilon^2\beta}{2}\mathcal{U}(\pi^*, \mathcal{F})$.
\end{theorem}

As compared with $\sigma^{1/m}$, we remove $1/m$ from the above approximation ratio, however, the time complexity of $\sigma^{\mathrm{DO}}$ could be exponential.
\section{Conclusion}
To the best of our knowledge, we are the first to systematically study the problem of adaptive robust optimization with nearly submodular structure. We analyze the adaptivity gap of ARONSS. Then we propose a approximate solution to this problem when all reward functions are submodular. Our algorithm achieves approximation ratio $(1-1/e)$ when considering matroid constraint.  At last, we develop two algorithms that achieve bounded approximation ratios for the general case.

\bibliographystyle{model1a-num-names}
\bibliography{reference}

\end{document}